\documentclass[conference]{IEEEtran}
\IEEEoverridecommandlockouts
\usepackage{cite}
\usepackage{amsmath,amssymb,amsfonts}
\usepackage{graphicx}
\usepackage{textcomp}
\usepackage{xcolor}
\usepackage{algpseudocode}
\usepackage{algorithm}
\usepackage{bbm}
\usepackage{siunitx}
\usepackage{subfigure}

\makeatletter
\def\ps@IEEEtitlepagestyle{%
  \def\@oddfoot{\mycopyrightnotice}%
  \def\@evenfoot{}%
}
\def\mycopyrightnotice{%
  {\footnotesize 979-8-3503-7550-3/24/$31.00$ $\copyright$ 2024 IEEE\hfill}
  \gdef\mycopyrightnotice{}
}
\def\BibTeX{{\rm B\kern-.05em{\sc i\kern-.025em b}\kern-.08em
    T\kern-.1667em\lower.7ex\hbox{E}\kern-.125emX}}
\begin{document}

\title{Maintenance Required: Updating and Extending Bootstrapped Human Activity Recognition Systems for Smart Homes \\
}

\newcommand{\thomas}[1]{{{\textcolor{blue}{[#1 -- Thomas]}}}}
\newcommand{\shruthi}[1]{{{\textcolor{blue}{[#1 -- Shruthi]}}}}
\newcommand{\edit}[1]{{\textcolor{black}{#1}}}
\newcommand{\foot}[1]{{\textcolor{red}{#1}}}

\author{\IEEEauthorblockN{Shruthi K. Hiremath}
\IEEEauthorblockA{\textit{School of Interactive Computing} \\
\textit{Georgia Institute of Technology}\\
Atlanta, USA \\
shiremath9@gatech.edu}
\and
\IEEEauthorblockN{Thomas Pl{\"o}tz}
\IEEEauthorblockA{\textit{School of Interactive Computing} \\
\textit{Georgia Institute of Technology}\\
Atlanta, USA \\
thomas.ploetz@gatech.edu}}

\maketitle

\begin{abstract}
Developing human activity recognition (HAR) systems for smart homes is not straightforward due to varied layouts of the homes and their personalized settings, as well as idiosyncratic behaviors of residents.
As such, off-the-shelf HAR systems are effective in limited capacity for an individual home, and HAR systems often need to be derived ``from scratch", which comes with substantial efforts and often is burdensome to the resident. 
Previous work has successfully targeted the initial phase.
At the end of this initial phase, we identify seed points. 
We build on bootstrapped HAR systems and introduce an effective updating and extension procedure for continuous improvement of HAR systems with the aim of keeping up with ever changing life circumstances. 
Our method makes use of the seed points identified at the end of the initial bootstrapping phase. 
A contrastive learning framework is trained using these seed points and labels obtained for the same.
This model is then used to improve the segmentation accuracy of the identified prominent activities.
Improvements in the activity recognition system through this procedure help model the majority of the routine activities in the smart home. 
We demonstrate the effectiveness of our procedure through experiments on the CASAS datasets that show the practical value of our approach.
\end{abstract}

\footnote{\normalsize{\foot{accepted at The 6th International Conference on 
Activity and Behavior Computing; under print @ IEEE Xplore. Contact shiremath9@gatech.edu for recent updates.}}}
\section{Introduction}
\label{sec:introduction}
\noindent
Developing robust and reliable human activity recognition systems for smart homes is essential, for example, to provide automated assistance to residents, or to longitudinally monitor daily activities for health and well-being assessments. 
\edit{With countries facing challenges in meeting elderly care requirements \cite{asadzadeh2022review,ahad2021activity}, such assessments can help in tracking behavior changes over a longer duration of time and prove as beneficial ambient assisted living (AAL) systems.}
To alleviate privacy concerns arising from camera-based monitoring and owing to advancements in IoT technologies the use of ambient sensors has been on the rise. 
Such sensing mechanisms provide for accessibility of reliable and inexpensive sensing and computing technology and instrumenting homes with sensors for everyday activity recognition in real-world living environments is now a realistic option for many. 
Although such advancements have made the data collection process seamless and straightforward, substantial challenges remain for developing and deploying HAR systems in smart homes \cite{bouchabou2021survey, bouchabou2021har, yadav2021review, benmansour2015human}.

Despite considerable progress in developing said activity recognition  systems \cite{liciotti2020sequential, bouchabou2021fully, bouchabou2021using, ghods2019activity2vec, aminikhanghahi2019enhancing,knox2010using}, drawbacks  exist.
Since smart homes are individualized settings with idiosyncratic behaviors, utilizing an ``off-the-shelf" activity recognition system is typically challenging and not straightforward.
Thus, a HAR system must be designed for individual smart homes, that caters to specific home layouts and activity patterns of its residents.
Such a system would require large amounts of data and annotations that can be used for building the fully-supervised and personalized models for a given home. 
In practical deployment scenarios, it is unreasonable to assume that a resident is willing to wait extended periods of time or provide extensive annotations until enough data is collected to develop a fully functional system. 
Also, for real-world deployments, privacy and logistical concerns essentially rule out that third parties will be able to collect the much needed annotated sample data while
the resident already lives in their smart home. 
As such, the focus is often on developing an initial model that provides for a limited recognition capability but is available ``early on" to the resident. 
These initial systems aim to capture prominent and frequently occurring activities \cite{hiremath2022bootstrapping}. 
Such, approaches are needed to derive functional HAR system quickly and with minimal yet targeted involvement of the residents themselves. 

\hspace*{4cm}\begin{figure*}[t]
    \centering
    \includegraphics[width=180mm,height=80mm]{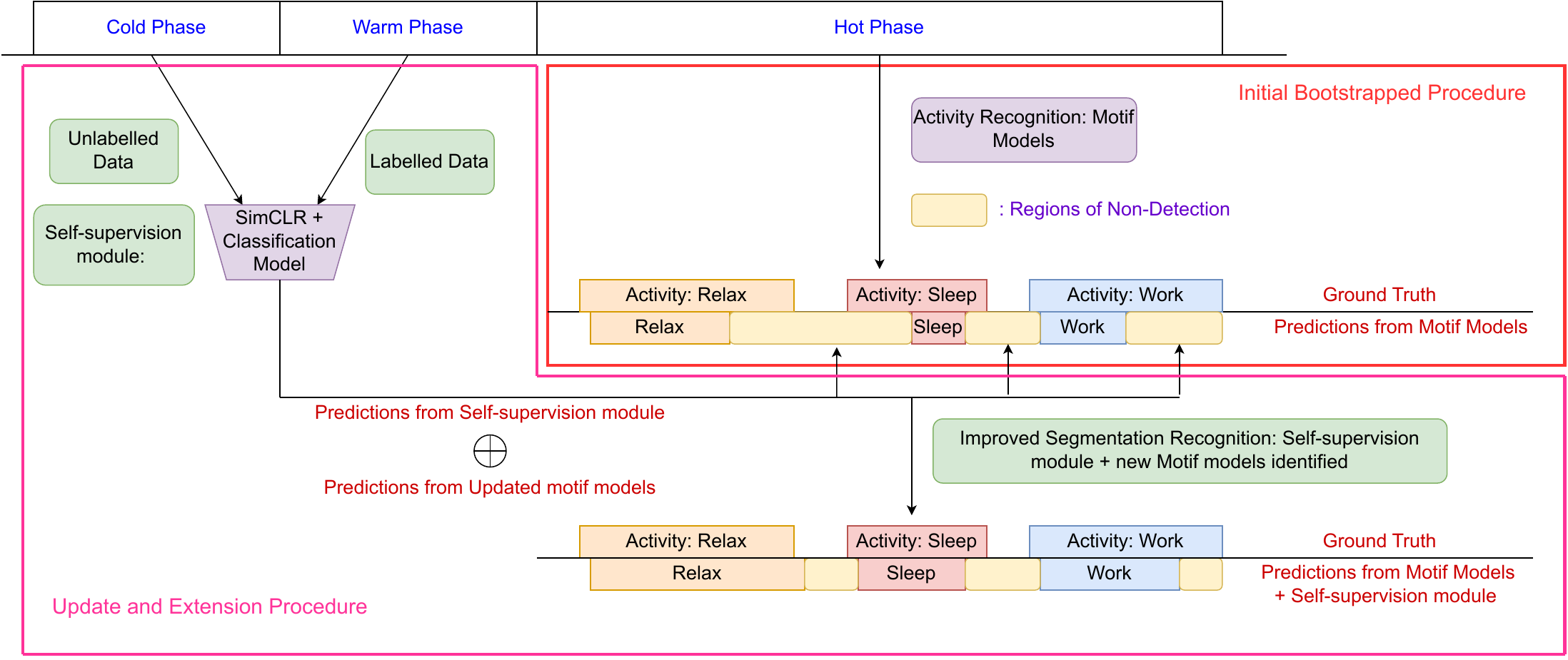}
    \caption{\edit{Updating activity models for HAR system.
    Activity predictions from the initial bootstrapped procedure (top-right portion) are used as starting points \cite{hiremath2022bootstrapping}.
    A self-supervision based module--SimCLR--utilized to learn representations is trained using unlabelled data, with sparse annotations from the active learning like procedure. 
    This module is then used to provide predictions in the non-detection regions produced through the motif models in the initial bootstrapping procedure. 
    Updated motif models are also learnt in a data incremental procedure.
    These modules make up the update and extension procedure (bottom portion)
    Predictions from both these  modules lead to improved segmentation accuracy for prominent activities that form majority of routine activities in the home.}}
    \label{fig:Fig1-Improved-Segmentation}
    \vspace*{-1em}
\end{figure*}

The initial system is used for coarse-grained recognition,
which will serve as the basis for further continuous and–again targeted–improvement and extension.
In this work, we develop a method that–based on this initial, functional yet not perfect version of a HAR system–extends the capabilities of an activity recognition system by integrating improved segmentation accuracy on the identified prominent activities. 
This maintenance and extension approach is possible through the `seed
points’ of activity recognition along a continuous monitoring timeline (for example over the span of a few weeks) provided by the initial system. 
We make use of the seed points identified by the initial model to train a self-supervision module.
Annotations provided in the minimal fashion by the residents are used to learn the mapping between the representations learnt from the self-supervision scheme and the labels corresponding to those seed points. 
A number of methods broadly belonging to the self-supervised paradigm, have made use of unlabeled data to learn robust data representations. 
Recent work in \cite{chen2023leveraging} has shown the use of SimCLR(simple framework for contrastive learning of representations) as an effective contrastive learning based self-supervision technique to learn the underlying representation space in smart homes. 
The objectives and contributions of this paper can be summarized as follows:
\begin{itemize}
\item \textit{Development of a Maintenance Procedure for the Activity Recognition System for Smart Homes --} \quad The update and extension procedure are used to improve the segmentation accuracy for recognized activity segments. 
This procedure is detailed in Fig. \ref{fig:Fig1-Improved-Segmentation}.
\item \textit{Evaluation of the developed Maintenance Procedure --}  \quad We demonstrate the effectiveness of our maintenance and extension approach through an extensive experimental evaluation on real-world smart home scenarios, namely within the context of the CASAS datasets \cite{cook2012casas}. 
Improvements in the activity recognition system are characterized by the majority of routine activities being covered well, i.e., exhibiting high classification as well as segmentation accuracy for the analysis of the prominent activities recorded through the continuous sensor readings. 
\item \textit{Application: Activity Logging --} \quad As such, our approach keeps track of a resident’s life and moves the HAR system closer to fully covering what is going on in a resident’s life.
Our contributions get us one step closer to developing an activity recognition system for smart homes in a fully data-driven manner.
\end{itemize}

\section{Related Work}
\label{sec:related-work}
\noindent
With decreasing sensor costs, automating `regular' homes has become a possibility for many. 
Such sensors can collect data for extended periods of time without concerns arising regarding 
battery re-charging (as in wearable sensors) or privacy (as in vision-based sensors). 
Analyzing such sensor data for human activity recognition purposes typically follows five steps \cite{bulling2014tutorial}:
\textit{i)} data capture; 
\textit{ii)} pre-processing to remove noise; 
\textit{iii)} segmenting the data stream into static data points that are assumed to be independent and identically distributed; 
\textit{iv)} feature extraction -- learning relevant information from data points; and 
\textit{v)} classification -- identifying the activity label for a given data point. 
\edit{The  approach proposed in this work integrates a set of techniques to continually update activity patterns as data is observed in the smart home. 
The initial activity patterns are obtained from previous work in \cite{hiremath2022bootstrapping}.
We use these as a starting point for our procedure (as described in Fig. \ref{fig:Fig1-Improved-Segmentation}) where the activity patterns, provide `seed points' to update activity segments corresponding to the prominent activities of interest.
We make use of a self-supervision based module by passively observing data in the smart home and build knowledge of representations incrementally. 
We also make use of updates to the recognition models from the initial bootstrapping procedure to learn changing activity patterns. 
In learning these representations, the goal is to improve the segmentation accuracy of the identified activity patterns. 
Thus the update and extension procedure is able to identify longer activity segments in comparison to previous work in \cite{hiremath2022bootstrapping}.
In what follows, we summarize the relevant related work  and the various components that are of relevance to our work.}

\vspace*{-0.5em}
\subsection{Activity Recognition in Smart Homes}
\label{sec:related-work:ar-smart-home}
\noindent
\edit{Activity recognition systems for smart homes typically aid in identifying instances of activities of daily living \cite{bouchabou2021survey, van2010activity, wongpatikaseree2012activity, rafferty2017activity}}.
This has a direct impact in terms of logging relevant behaviors in monitoring health-related scenarios and in identifying changes that might occur from regular routines.
In order to track the resident's activity or behavior, data is collected through networked devices  in the home. 
Ambient sensors that record event-based data are employed to record state changes.
\edit{These sensor events are recorded in a continuous fashion, resulting in a time-series problem.}
However, data collected through ambient sensing mechanisms does not have a continuous sampling rate as opposed to data collected through either wearable sensors or videos \cite{hussain2019different, chen2012sensor}.
A number of algorithms have been proposed for the task of sensor-based human activity recognition systems that use K-nearest neighbors, random forests \cite{sedky2018evaluating}, Hidden Markov Models, and support vector machines \cite{cook2010learning}. 
More recent work in \cite{liciotti2020sequential} explores using sequential models (and variants thereof) in modeling activities of interest.
\edit{In \cite{bouchabou2021fully, bouchabou2021using} various language-based encodings, such as -- ELMO and Word2Vec, are used to represent the data observed.}
However, most, if not all, contemporary works use the `segment-first and then recognize' processing approach.
\edit{Such procedures are not ideal for real-world deployments where the resident would need to provide activity start and end points.}
Based on the smart home layout and the activities of interest, various change point detection (CPD) algorithms have been applied to segment activities \cite{aminikhanghahi2018real,sprint2020behavioral,jose2017improving}. 
These procedures identify abrupt changes in the sensor data streams by employing a heuristic measure based on the statistical properties of the signal or likelihood ratios \cite{aminikhanghahi2017using, aminikhanghahi2018real}. 

Approaches that make use of either time-based windowing or (sensor) event-based windowing, that relax the requirement to know the start and end times of an activity in the `segment-first and then recognize' approach have been employed. 
Features learnt over these windows are either \textit{i)} handcrafted; or \textit{ii)} learnt automatically. 
Hand-crafted features encode information such as time of the last sensor event in the window, day of the week corresponding to the last sensor event in the window, dominant sensor in the window, last sensor location in the window to name a few. 
\edit{Automatically learned features are extracted, to name a few procedures, through employing  \textit{i)} sequential-modeling procedures \cite{ghods2019activity2vec}; \textit{ii)} graph-based approaches \cite{zhou2020graph, li2019relation}; \textit{iii)} self-supervision based approaches \cite{chen2024enhancing,chen2023leveraging }; \textit{iv)} convolution-based approaches over sensor data represented in the form of images \cite{gochoo2018unobtrusive,mohmed2020employing,singh2017convolutional}. } 

Although, the aforementioned approaches report reasonable scores on the activity recognition task, the metric often used for evaluating the built systems is the--misguiding--F1-weighted score.
The use of this metric falsely suggests satisfactory performance of the analysis procedure, since it weighs the activity class that occurs in majority more than those that occur less frequently.
However, in the context of the CASAS datasets used for analysis, due to an imbalance in the datasets, the `Other' class, which is an activity class not of interest occurs more frequently than activities of interest such as `Enter Home'. 
Thus, providing more weight based on the number of instances of an activity biases the classification towards predicting the majority class most of the time, without learning much about the actual activity classes of interest.

\subsection{Self-supervision based representational learning}
\label{sec:related-work:representation learning}
\noindent
\edit{The goal of this work is to continually improve the segmentation of the activities identified through the bootstrapping procedure.
In order to continually update representations in an unsupervised fashion, we focus on learning these representations without the need for labels.} 
Self-supervised learning is a powerful technique that allows for extracting feature representations from large amounts of unlabelled data. 
The self-supervision procedure consists of two stages: \textit{i)} designing the pretext task for learning robust feature representations; and \textit{ii)} the fine-tuning procedure on downstream tasks aimed at transferring the knowledge learnt from the pretext task to specific tasks by fine-tuning the features. 
Broadly, there are two approaches to self-supervised learning: \textit{i)} Contrastive learning: where the aim is to distinguish between similar and dissimilar data points within the input data; and \textit{ii)} Non-contrastive learning: where only positive samples are used to learn the feature representations -- the model is aimed at bringing the original data point and its augmented version close. 
No negative data points are used during the learning process. 

SimCLR \cite{chen2020simple}, a popular contrastive learning approach that makes use of three major components to learn good representations:
\textit{i)} using appropriate data augmentations; 
\textit{ii)} making use of a non-linear transformation layer also known as a projection head; and
\textit{iii)} optimizing the contrastive learning loss over larger batch sizes and training steps. 
This learning procedure does not require a memory bank or specialized architectures in order to learn the useful representations. 
SimCLR has been extensively used in time-series analysis problems and has shown promising results \cite{yang2022timeclr,shah2021evaluating,mohsenvand2020contrastive,haresamudram2022assessing}. 
In recent works \cite{chen2023leveraging}, the usefulness of utilizing a modified version of SimCLR in ambient settings has shown promising results. 
Unlike using raw data samples to learn robust representations, features are extracted from raw ambient data which are then utilized to learn the representation space. 
The encoder consists of two layers of CNN followed by one layer of LSTM and the projection head consists of three fully connected layers.

\subsection{Active Learning}
\label{sec:related-work:active learning}
\noindent
Developing large-scale activity recognition systems usually requires large amounts of labeled instances of data. 
To reduce such a reliance on annotation as a resource \cite{hiremath2020deriving}, a semi-supervised learning paradigm--Active Learning--uses a human in the loop to obtain annotations \cite{settles2009active}.
By providing annotations for limited amounts of relevant and informative data points, the goal is to reduce the requirement of large amounts of annotations and provide comparable performance scores to the fully supervised conditions. 

The Active Learning paradigm has two components:
\textit{i)} the sampling strategy, which details the schedule for data points to be picked for annotations from the unlabeled dataset; and  \textit{ii)} the query strategy, which defines a heuristic function that determines the data points to be labeled based on the evaluation of the heuristic function.  

Pool based active learning is employed for tasks where large samples of unlabeled data are available. 
A classifier is first trained on a small labeled dataset \cite{hossain2017active, adaimi2019leveraging, jones2014active} and a query strategy utilizing the already trained classifier then identifies data points to be queried from the pool of unlabeled dataset for further annotation.
Stream-based active learning is beneficial in online settings where pool-based strategies cannot be applied. 
A budget is usually defined to identify the number of queries that can be made in the active learning procedure. 
This is generally limited and tuned based on the performance expected from the system and the burden a human annotator is expected to accept. 
Different budget spending strategies have been explored in \cite{miu2014strategies}.

\edit{In the bootstrapping procedure, annotations corresponding to the identified motifs are obtained from the residents in an active learning like procedure. 
We make use of these annotations to train the self-supervision module.
Thus the reliance on the resident to provide annotations is kept at a minimum throughout the entire procedure. 
For the updates to the recognition procedure, through the activity models, the procedure (similar to the initial bootstrapping procedure) of asking for minimal labels from residents every few weeks is retained.}

\section{Updating and Extending Bootstrapped Human Activity Recognition Systems for Smart Homes}
\label{sec:sec3-intro}
\noindent
In this work, we develop a procedure to update HAR systems in continuously evolving smart home environments. 
We focus on utilizing the `seed points' of the initial bootstrapping procedure as the starting point for analysis. 
The `seed points' provide for initial segments of activities identified through the activity recognition procedure in the initial bootstrapping stage.
The identified segments are aimed at discovering the occurrence of the activity. 
As such these segments may not cover the entire length of the activity as it occurs in the smart home. 
Our `update and extension' procedure is aimed at extending these seed points and bringing the activity recognition system closer to identifying the correct start and end points of the activities in the smart home. 
We employ a conjunction of a self-supervision module and an updated activity models in order to update the recognition system. 
A data-incremental procedure is used to update the recognition model, where data collected over every few weeks in the smart home is used train the recognition procedure.   

\subsection{Scope} 
\label{sec:sec3-scope}
\noindent
We develop an update procedure that aims to extend patterns of already identified activities through the initial bootstrapping procedure, and to learn new patterns corresponding to activities the system was unable to identify previously -- the HAR model will be updated and extended. 
For this update procedure, we make some assumptions that help scope out the contribution of this work. 

\edit{We use the initial bootstrapping procedure in \cite{hiremath2022bootstrapping} as the starting point and obtain labels corresponding to \textit{newly} identified motif models from the resident through an active learning like procedure \cite{ciliberto2019wlcsslearn}.}
The proposed procedure assumes that the predictions of the bootstrapping procedure are accurate with regard to the predicted activities but they may not be precise with regard to the actual boundaries of the detected activities, i.e., while the predictions will match with activities, the actual activities might extend those seed points.

Since we utilize \cite{hiremath2022bootstrapping} as the starting point for the work presented here, we retain the design choices used in the initial work. 
The design choices used in the previous work are:
\textit{i)} the length of the action unit is determined through observing 20 sensor events, and co-occurrence patterns across these sensor event triggers are learned through the embedding layer; 
\textit{ii)} motifs for the prominent activities have to be $\geq$ 2 in length; 
\textit{iii)} motifs for prominent activities should occur $\geq$ 5 times; 
\textit{iv)} motifs identified should be homogeneous in the activity label; and
\textit{v)} motifs for prominent activities are identified through majority voting to ensure these models are strong predictors of a given activity.

\begin{figure*}
    \centering
    \includegraphics[width=0.8\textwidth,height=80mm]{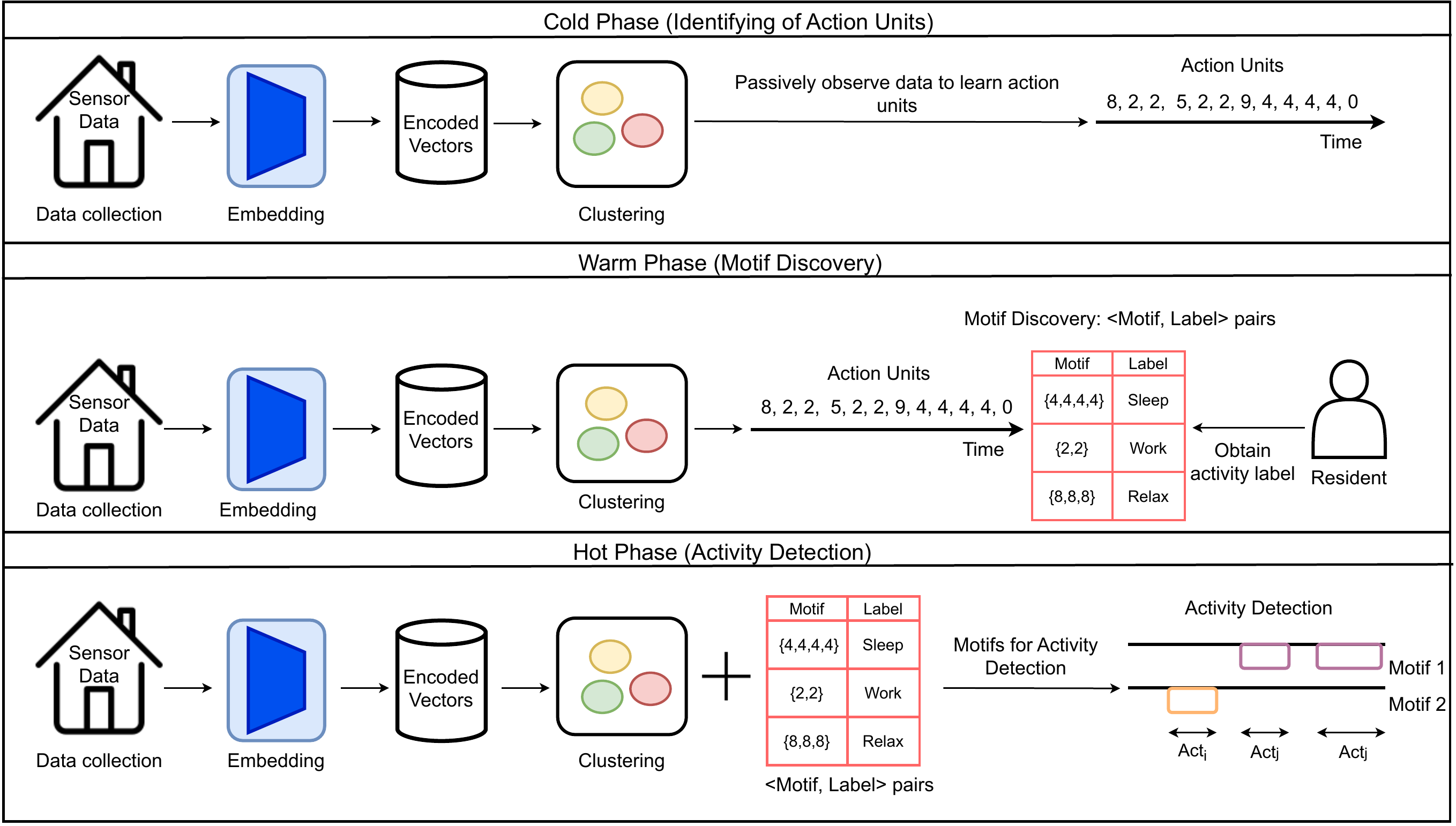}
     \caption{\edit{Summary of the initial bootstrapping procedure for Phase 1 of the HAR system lifespan (taken with permission from \cite{hiremath2022bootstrapping})}}    \label{fig:bootstrapping}
\end{figure*}

Both the initial bootstrapping and update procedures aim to provide solutions through a data-driven approach, without requiring too much effort from the resident's end. 
Predictions from both these procedures can be refined by utilizing additional information such as an ontology built for the smart home identifying relevant sensor events corresponding to a given activity or using knowledge of activity routines that the resident engages in. 

\edit{The update procedure starts after the initial \textit{n+m} weeks of data in the smart home, wherein the first \textit{n} weeks are utilized during the \textit{Cold Phase} and the next \textit{m} weeks during the \textit{Warm Phase} as shown in Fig. \ref{fig:bootstrapping}.
Previous work in \cite{hiremath2022bootstrapping} uses \textit{n} and \textit{m} equal to two weeks each. 
\ref{fig:bootstrapping}
However this is a hyper-parameter that can be varied.
We hypothesize that the  procedure, after the required break-in period, provides a good starting point for the update and extend procedure.} 

\subsection{Prerequisites: Initial Bootstrapping Procedure}
\label{sec:sec3-prereq}
\vspace*{-0.5em}
\noindent
\edit{Action units are learned through an embedding and clustering procedure that correspond to the movement patterns in the home, learnt during the \textit{Cold Phase} (top portion of Fig. \ref{fig:bootstrapping}).
Algorithm \ref{alg:predictAU} details the procedure to obtain these action units that serve as building blocks for the next steps of the bootstrapping procedure.
The action units are learnt over the first two weeks (\textit{n}=2) of data observed in the home. 
In the second stage-–\textit{Warm Phase} (middle portion of Fig. \ref{fig:bootstrapping})--frequently occurring sub-sequences comprising of action units are identified through a set of filtering procedures and queried. 
Activity labels corresponding to these sub-sequences are collected through the query procedure from the resident. 
Thus, a set of motif models, which represent activity models are derived for a sub-set of activities of interest. 
A merge or overlap is used to merge motif models of varying lengths to obtain the final models. 
The length of these final motif models provides initial segmentation boundaries for the detected activity.}

\edit{Motif models are learnt over the next two weeks (\textit{m}=2) (after the \textit{Cold Phase}), of data observed in the home. 
During the last stage-–\textit{Hot Phase} (bottom portion of Fig. \ref{fig:bootstrapping})--the system is deployed to detect activities occurring in the smart home.
After the first \textit{n+m} weeks (\textit{n+m}=4)  required for the initial bootstrapping procedure, the initial set of motif models is available.}

\subsection{Self-supervision module}
\label{sec:sec3-self-supervision}
\noindent
\begin{figure}[t]
    \centering
    \includegraphics[width=0.5\textwidth,height=50mm]{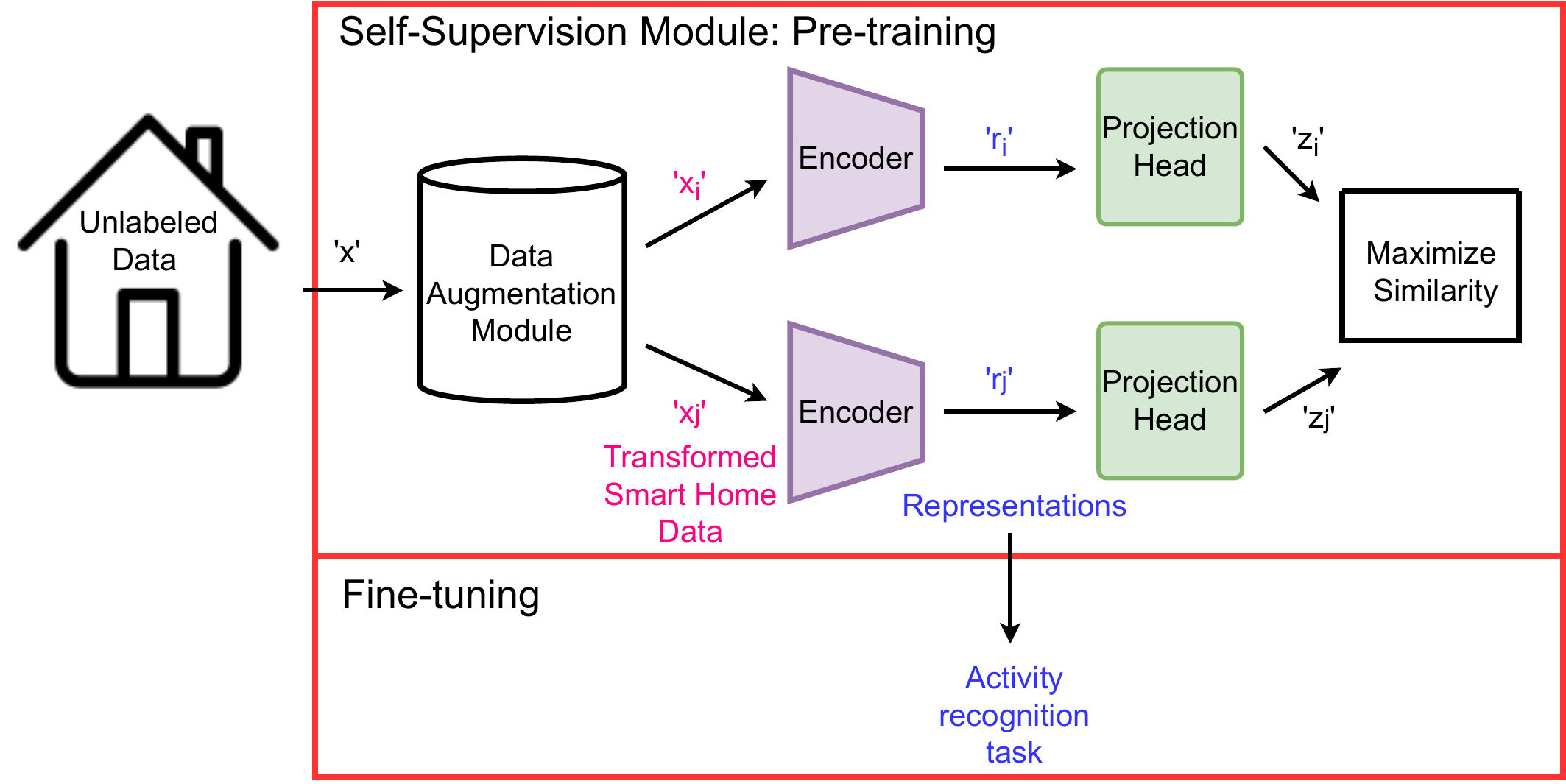}
    \caption{\edit{Architecture for the self-supervision module. The pre-tranining module (top) requires only unlabeled data for training. The fine-tuning module (bottom) makes use of representations from the self-supervision module and annotations to train a model for an activity recognition task. See text in Sec. \ref{sec:sec3-self-supervision} for details of the architecture.}}     
    \label{fig:simclr}
    \vspace*{-1.5em}
\end{figure}
We employ self-supervision to extend the `seed points' obtained during the bootstrapping procedure.
The self-supervision module serves to learn good representations from data observed in the smart home.
Since the self-supervision technique learns discriminative and meaningful representations from unlabeled data, the resident in the loop is not burdened with providing annotations for each data point observed in the home. 
\edit{The components of the self-supervision module are illustrated in Fig. 3. }
This module consists of two components: \textit{i)} pre-training; and \textit{ii)} fine-tuning.

In the pre-training procedure representative features are learnt from large amounts of unlabeled data. 
We detail the components of the self-supervision module below: 
\begin{enumerate}
\item Representations are learnt over the embeddings of the actions units. 
These embeddings are learnt from the BERT model \cite{hiremath2022bootstrapping}, which uses a 15\% masking probability to obtain the encodings over which the action units are obtained.
\edit{These representations are fed into the self-supervision module and are denoted as `$x$' in Fig. \ref{fig:simclr}.}
\item We make use of the noise transformed and scaling transformed data augmentation techniques to obtain augmented pairs of the input data. 
\edit{These augmented pairs are represented as transformed datapoints `$x_i$' and `$x_j$', where `$x$' undergoes the noise transformed and scaling transformations.}
A random noise is applied to the data point for the noise based transformation and scaled by a normal distribution for the scaling based transformation. 
These transformations were introduced in \cite{tang2020exploring} and have proven effective for time series data.
\item The encoder consists of two convolutional layers, followed by an LSTM layer. 
The filter sizes for the convolutional layer are 32 and 64. 
Each convolutional layer is followed by a RELU and dropout layer (with dropout probability of 0.1). 
The number of hidden units for the LSTM layer are 64.
\edit{Transformed data points `$x_i$' and `$x_j$' are passed through this encoder to obtain embeddings as `$r_i$' and `$r_j$' respectively.}
\item \edit{The embeddings obtained (`$r_i$' and `$r_j$') are then passed onto the projection head, which consists of three linear layers with input filter sizes of 64, 128, and 256. 
Every linear layer is separated by a RELU layer.
This gives us new embeddings--`$z_i$' and `$z_j$'--which are then used as inputs to the loss function.}
\item A contrastive loss function, NT-Xent \cite{jiang2020robust}, is used to maximize the similarity between augmented data points and minimize the similarity to the other data points in a given batch. 
The loss function is defined in Eq. \ref{eq:nt-xent}.
\edit{`$z_i$' and `$z_j$' correspond to the representations obtained from the self-supervision module for an unlabeled data point `$x$' collected in the smart home. 
`$z_i$' and `$z_j$' correspond to positive pairs in the batch whereas `$z_i$' and `$z_k$' correspond to the negative pairs.
A batch size of 64 is used.
`$k$' represents the number of data point(s) in a given batch.
Since each unlabeled data point results in 2 augmented versions the number of datapoints is two times the pre-defined batch size (2N).
The cosine similarity is scaled by a temperature parameter $\tau$.
}
\begin{equation}
\label{eq:nt-xent}
    L_{i,j} = -log \dfrac{exp(sim(z_{i}, z_{j})/ \tau}{\sum_{k=1}^{2N} \mathbbm{1}_{[k \neq i]}exp(sim(z_{i}, z_{k})/ \tau}
\end{equation}
\end{enumerate}

In the fine-tuning procedure, we make use of small amounts of labeled data to fine-tune the encoder pretrained in the previous step. 
The data points and corresponding labels are used from the predictions of the bootstrapping procedure. 
Since the bootstrapping procedure identified prominent activities, the self-supervision module learns a model corresponding to only these prominent activities, with the goal of improving segmentation accuracy. 
The projection head from the pre-training step is discarded and only the encoder is used to obtain the embeddings.
Similar to \cite{chen2023leveraging}, the prediction head consists of two sequential layers with input feature sizes of 64 and 256. 
A RELU layer is used for activation between the two sequential layers. 
Cross-entropy loss is used for the activity recognition task.

\vspace*{-0.5em}
\subsection{Update and Extend the Initial Bootstrapped Procedure}
\label{sec:sec3-update-extend}
\noindent
In this work, we design the update and extend procedure (Fig.\ \ref{fig:Fig1-Improved-Segmentation}) of the activity recognition system.
The update procedure starts after the initial bootstrapping procedure (Fig.\ \ref{fig:bootstrapping}), which serves as a pre-requisite and starting point for the update procedure. 

The motif models obtained provide the initial segmentation corresponding to the prominent activities in the home.
Although these segments provide a good initialization point to identify the activities of interest, they are not accurate in providing the accurate start and end points corresponding to these activities.
In the update and extend procedure, we use these initially identified segments to train the self-supervision module. 
The predictions are used as labels to fine-tune the self-supervision module.
Softmax score obtained from the trained model are used to predict the occurrence (or not-) of the identified prominent activities. 

In parallel, the motif discovery procedure identifies (new-) motifs with every update. 
Since the data observed during the \textit{Cold Phase} increases with each update, new motifs capture increasing number of movement patterns. 
The predictions from the self-supervision model and the new motifs discovered are combined to obtain the final segmentation accuracy score.

\begin{figure}[t]
    \vspace*{-1em}  
    \hspace*{2em}
    \includegraphics[width=0.3\textwidth, height=35mm]{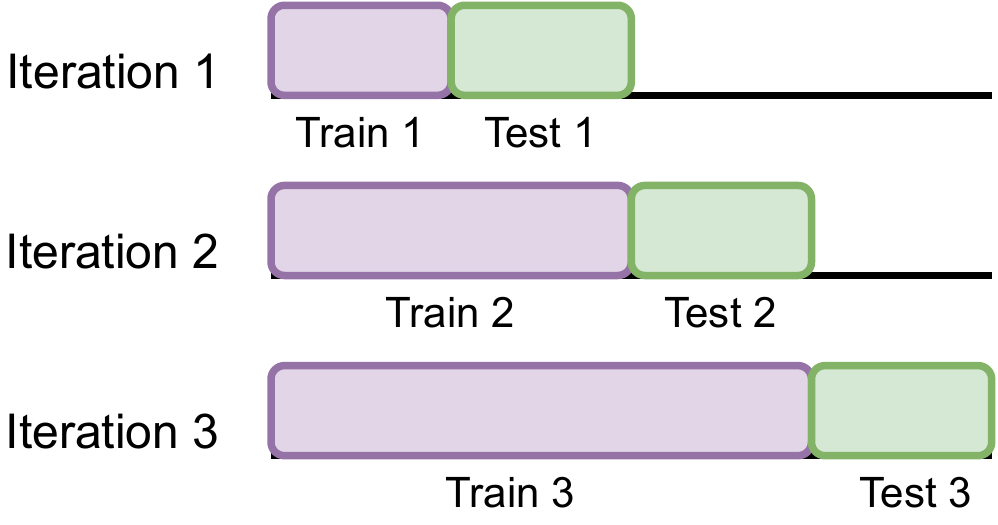}
    \caption{\edit{Data incremental procedure for update and extension of the bootstrapped model.}}   
    \label{fig:fc}
\end{figure}

\begin{algorithm}[t]
\small
\caption{Cold Phase: Predict Action Units (with permission from \cite{hiremath2022bootstrapping})}
\label{alg:predictAU}
\edit{\begin{algorithmic}
\State \textbf{Input:} Data D: \{$\langle sensor_j, value_j\rangle\}$; BERT; k-Means 
\Comment {BERT and k-Means are trained models as described in Sec. \ref{sec:sec3-prereq}. Input data stream is made up of sensor event triggers ($sensor_j$) and their corresponding values ($value_j$)}
\State \textbf{Output:} Action Unit Predictions AU: \{$au_1, au_2, ...,au_i$\}
\State windows = sliding\_windows(D)
\For {$ w_i \in  windows $} 
\State $encoding_i$ = BERT($w_i$)
\State $au_i$ = k-Means($encoding_i$)
\State $au_i$ $\rightarrow$ AU 
\EndFor
\end{algorithmic}}
\end{algorithm}

\begin{algorithm}[t]
\small
\caption{Update \& Extend Initial Bootstrapped Procedure}
\label{alg:update}
\edit{\begin{algorithmic}
\State \textbf{Input:} $M_{t} = \{M_1, M_2, M_3, ..M_j\}$; 
\Comment{Initial Motif models ($M_{t}$)}
\newline Data $D_{t}$: \{$\langle sensor_j, value_j\rangle\}$; 
\Comment {Input data stream} 
\newline $\theta$: self-supervision model
\State \textbf{Output:} Updated activity segments  $A_{updated}$: $A_{t-1}$ + $A_{t}$ 
\Comment{$A_{t-1}$: refer Sec. \ref{sec:sec3-prereq}; $A_{t}$: refer Sec. \ref{sec:sec3-update-extend}}
\State windows = sliding\_windows(D)
\State {$AU = \textbf{predict action units} (windows)$}
\Comment {refer Alg. \ref{alg:predictAU}}
\For {$m_i \in M_{t}$}
\Comment{refer Sec. \ref{sec:sec3-prereq}}
\If{$m_i$ matches $AU[k:n]$}
\State $detection = detection + AU[k:n]$
\newline
\Comment {k $<$ n: AU sequence indices}
\EndIf
\EndFor
\State $non\_detection = AU - detection $
\State $A_{t}$  = $\theta (non\_detection) + M_{updated}(D_t)$ 
\Comment{ $\theta (non\_detection)$: refer Sec. \ref{sec:sec3-self-supervision}; $M_{updated}(D_t)$: refer Sec. \ref{sec:sec3-update-extend}}
\State  $A_{updated}$: $A_{t-1}$ + $A_{t}$ 
\State \textbf{return} $\langle A_{updated}\rangle$
\end{algorithmic}}
\end{algorithm}

\vspace*{-0.5em}
\subsection{Deployment and Activity Recognition }
\label{sec:sec3-deployment}
\noindent
We use a data incremental procedure during the deployment of the designed activity recognition system, where the activity recognition is updated every  $(n)$ weeks, through training the self-supervision module as shown in Fig. \ref{fig:fc}.
\edit{During the deployment procedure, the motif models stored in the motif memory up to the last time step $(t-1)$ are utilized. 
The first $(n+m)$ weeks are from initial bootstrapped procedure and the next $((t-1)*n)$ weeks are from the data observed in update and extend procedure.}

An activity prediction is reported when there is a match between an activity model and a sequence of observations in the smart home. 
By utilizing the incremental versions through the update and extend procedure, the segmentation accuracy improves and the system gets closer to a fully-functional event-based recognition system. 
With increasing observation period in the home, large quantities of data become available to train the self-supervision module.
\edit{The goal of building such update and extension mechanisms is to detect activity occurrences, which can provide activity logs that are to be used for activity monitoring and behavior analysis.}

\section{Experiments}
\label{sec:results}
\noindent
Through our experimental analysis we explore the effectiveness of our update and extension procedure, with specific focus on segmentation of  prominent activities identified in the bootstrapping procedure. 
The initial HAR system derived through the initial bootstrapping procedure;  illustrated in Fig.\ \ref{fig:bootstrapping} is developed for a given smart home using the first \textit{n+m} weeks of observations. 
We extend this procedure to now improve the segmentation accuracy for the prominent activities. 

As such, the update procedure is initiated after the first \textit{n+m} weeks of a resident living in the smart home.
Thus, the starting point for the maintenance procedure is through the previously defined \textit{Hot Phase}. 
\edit{Updates to the recognition model
are scheduled after every two weeks of data observed in the home. 
The number of weeks required for the update is a 
design choice that can be tweaked based on the requirements of the application.
However, since the initial bootstrapping procedure used increments of two weeks to develop a model, we retain this design choice during the update and extension procedure.
Thus, multiple versions of updated HAR systems are derived through observing--and processing blocks of \emph{n} consecutive weeks of data. 
This data incremental procedure is illustrated in Fig. \ref{fig:fc}.}

The evaluation procedure in this application scenario is challenging.
Ground truth is obtained retrospectively from residents through surveys and sensor event triggers are then analyzed to provide activity labels and boundaries. 
Such annotations provide for noisy ground truth and obtaining sample-precise evaluations proves challenging. 
The methods developed here can be used ``as-is" in any smart home, irrespective of varying layouts and residents' activity patterns. 
Since the update procedure aims at improving the segmentation procedure by extending the seed points from the bootstrapping procedure it brings us closer to providing a functional system that determines activities in the home.

\subsection{Methodology}
\label{sec:results-methodology}
\noindent
The update and maintenance procedure targets the improvement of the segmentation of the prominent activities identified in the bootstrapping procedure.
We provide experimental evaluations on the aforementioned CASAS datasets.
A continuous evaluation protocol is used wherein the self-supervision module is continuously updated and used to improve the segmentation accuracy.  
We provide quantitative results as part of the evaluation procedure and compare improvements  over the initial bootstrapping phase.

\textit{Evaluation Protocol:}
The self-supervision module is initially trained on the predictions from the bootstrapping procedure identified after the first \textit{$n+m$} weeks.
The trained recognition model is then evaluated on all data henceforth.
The same procedure is repeated after every \textit{n} weeks during the update phase, wherein once the self-supervision model is updated--at time step (\textit{$t-1$})--the recognition model is used for evaluation on all data observed in the home from the next (\textit{$n$}) weeks. 
Hence the updates to the self-supervision model come from \textit {$n+m+(t-1)*n$} weeks.

\vspace*{0.5em}
\subsection{Datasets and Data Pre-Processing}
\label{sec:results-datasets}
\noindent

\begin{figure}[t]
    \vspace*{-1em}
    \subfigure[CASAS-Aruba]
    { \hspace*{0.8cm}  
    \centering
        \includegraphics[keepaspectratio, width=.35\textwidth]{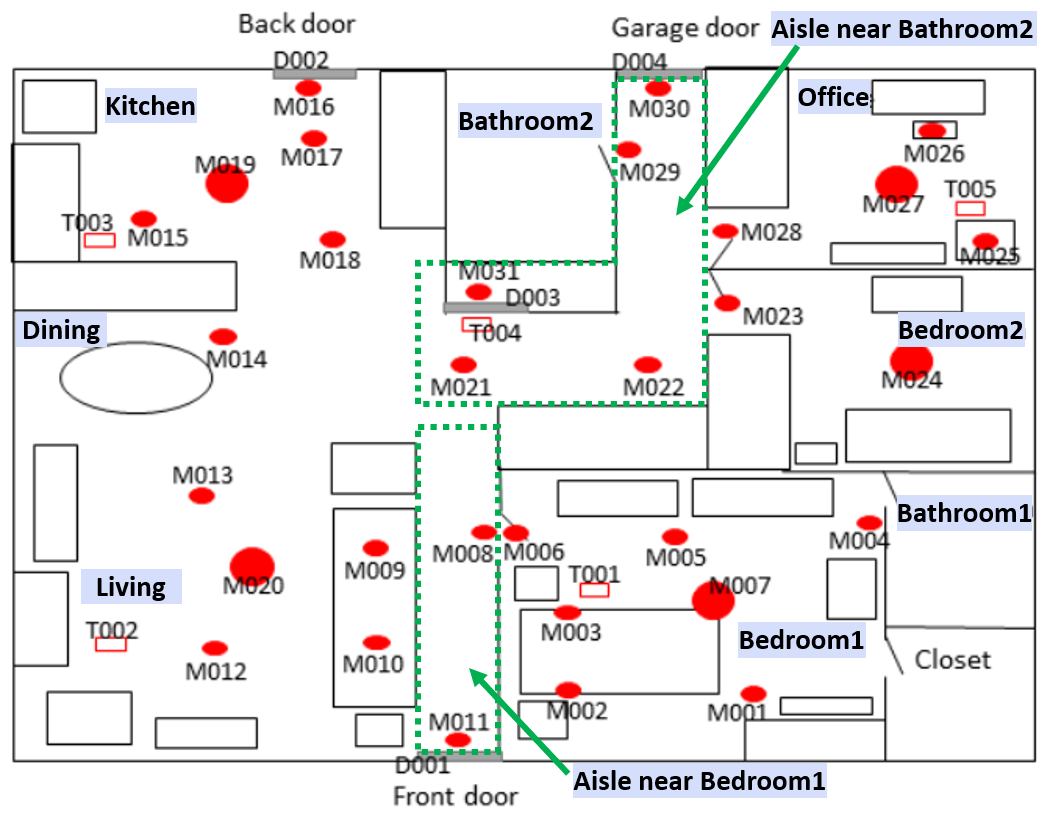}
       \label{fig:aruba}
    }
     \subfigure[CASAS-Milan]
    { 
     \hspace*{0.8cm}  
    \centering
        \includegraphics[keepaspectratio, width=.35\textwidth]{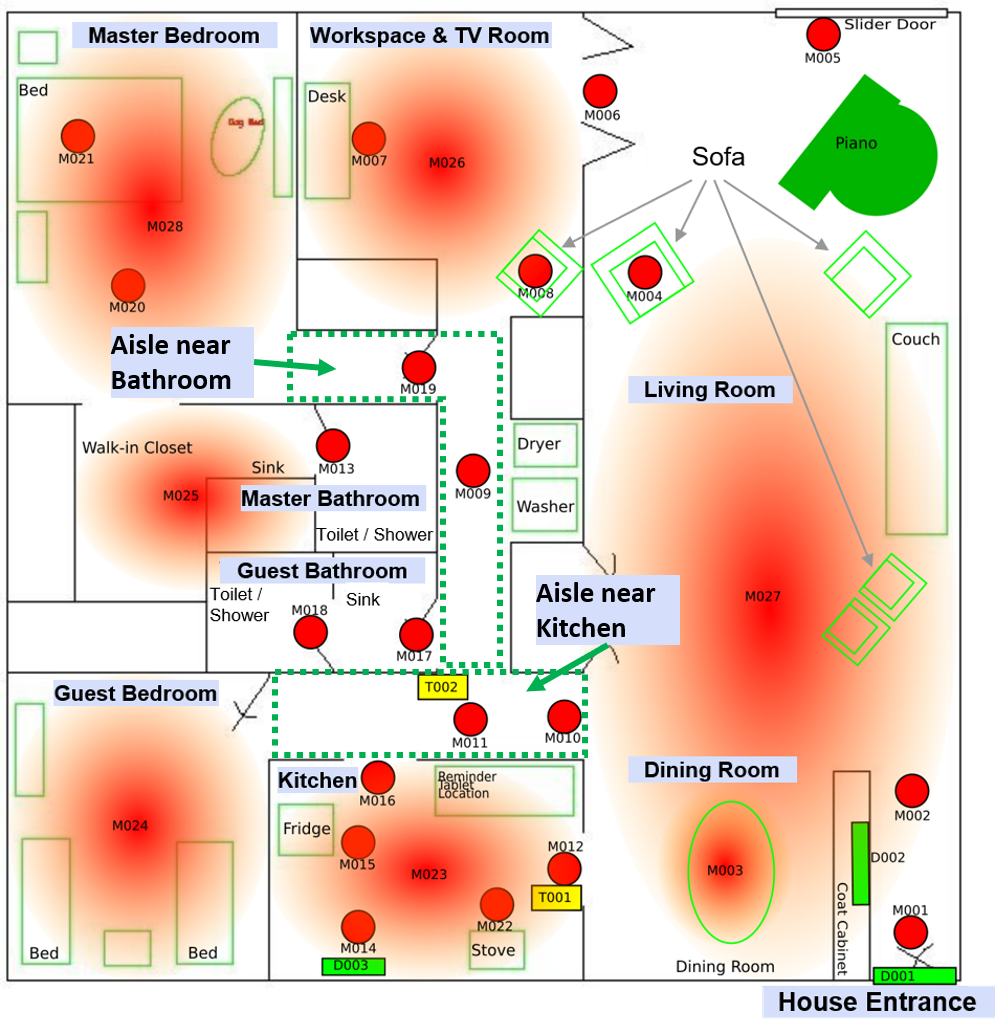}
        \label{fig:milan}
    }\\
    \subfigure[CASAS-Cairo]
    { 
     \hspace*{0.8cm}  
    \centering
        \includegraphics[keepaspectratio, width=.35\textwidth]{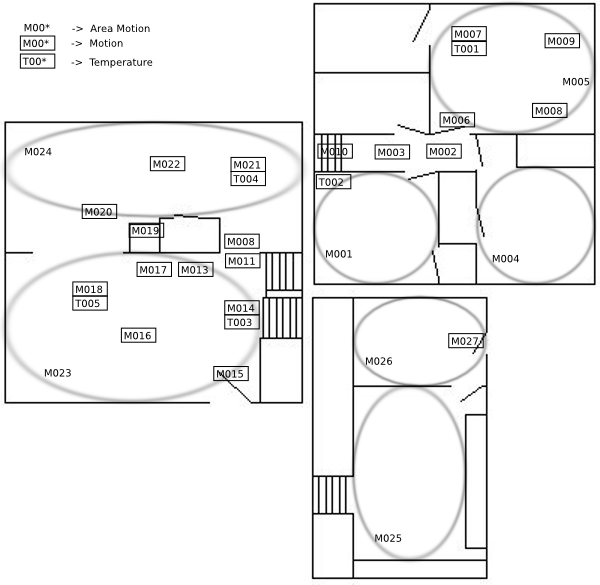}
        \label{fig:cairo}
    }
    \label{fig:layout}
    \edit{\caption{Floor plans for Smart Homes used for our experimental evaluation: (a) CASAS-Aruba, (b) CASAS-Milan and (c) CASAS-Cairo (with permission from \cite{cook2012casas}). Annotations for locations are used with permission from \cite{hiremath2022bootstrapping}.}}
    \vspace*{-1em}
\end{figure}

We base our explorations on publicly available datasets that are widely used in the research community.
These datasets were collected as part of the Center of Advanced Studies in Adaptive Systems (CASAS), with ground truth annotations provided by residents.

We evaluate the effects of our method on three popular datasets: CASAS-Aruba, CASAS-Milan and CASAS-Cairo, which were collected over 219, 92 and 56 days respectively.
CASAS-Aruba and CASAS-Milan are single-resident households, whereas CASAS-Cairo houses two residents. 
Both CASAS-Milan and CASAS-Cairo also house a pet.
\edit{The CASAS-Cairo home has three storeys, the detailed layout of each are depicted in Fig. \ref{fig:cairo}.}
Details of these datasets are in Tab. \ref{tab:datasets}.

\edit{Ambient sensors such as Motion (denoted as `M\#\#\#'), Door (denoted as `D\#\#\#') and Temperature (denoted as `T\#\#\#')  sensors are used to collect data in these homes  as shown in Fig. 5. 
The Door and Motion sensors are binary sensors, where the states of the Door sensors correspond to OPEN or CLOSE and that of Motion sensor corresponds to ON or OFF, respectively.
The sensors are numbered randomly, with no specific ordering. 
Door sensors, represented as green lines, specifically capture the opening and closing of the doors and drawers in the home (for example `D001' near the entrance of the home in CASAS-Milan) whereas Motion sensors represented by the red dots capture movement close to where they are located. 
The red dots in the smart home layouts represent the detection of motion in localized areas whereas the radiating red dots capture movement over a wider area.
Temperature sensors record changes in the home, but do not capture movement. 
Hence, as in previous work \cite{gupta2020tracking, hiremath2022bootstrapping}, we do not make use of these sensors.
We also do not make use of absolute timestamps to make our proposed approaches generalizable to implement across various smart home layouts and idiosyncrasies of the residents occupying them. }

\begin{table}
    \centering
    \vspace*{0.8em}
    \caption{Details of the CASAS datasets used for our experimental evaluation}
    \label{tab:datasets}
    \small
    \begin{tabular}{c||c||c||c} 
        CASAS Dataset & Aruba & Milan & Cairo \\
        \hline
        Days & 219 & 92 & 56 \\ [10pt]
        \hline
        Residents & 1 & 1+pet & 2+pet \\ [10pt]
        \hline
        Sensors & 39 & 33 & 27  \\ [10pt]
        \hline
         Activities & 11 & 14 & 6  \\ [10pt]
        \hline
        \end{tabular}
\end{table}

\subsection{Deployment and Activity Recognition}
\label{sec:results-ssl}
\noindent
Our update procedure is initiated after the first four weeks of the resident living in the home, which is when the initial HAR system was bootstrapped \cite{hiremath2022bootstrapping}, and it is then subsequently updated every two weeks. 
We evaluate the effectiveness of our update and maintenance by reporting the segmentation accuracy using ground truth labels provided for CASAS-Aruba, CASAS-Milan and CASAS-Cairo.
For the initial bootstrapping procedure, the derived HAR system was evaluated on all data of the \textit{Hot Phase}. 
For the update and extend procedure, we evaluate the updated HAR system continuously, i.e., after each block of two weeks as shown in Fig.\ \ref{fig:fc}. 
The results on segmentation accuracy are tabulated in and Tab.\ \ref{tab:results-ar-cairo}, Tab.\ \ref{tab:results-ar-aruba} and Tab.\ \ref{tab:results-ar-milan}. 
Each column in these tables represents the scores obtained from an update to the motif memory at time step \textit{$t-1$} for all data from time step \textit{$t$} until data is collected in the home. 
When no such update is observed, the corresponding block is empty.

\textit{Evaluation:}
\label{sec: Experiments-ar-quant}
The initial self supervision model is trained using the predictions from motif models obtained in the bootstrapping procedure. 
Since the bootstrapping procedure only recognizes prominent activities (a subset of the activities occurring in the home), the self-supervision module is trained on only these activities.
The goal of this work is to improve the segmentation accuracy corresponding to these prominent activities, that form essential components of daily routines.
The improved segmentation procedure will enable identification of the less frequently occurring activities by analyzing identified gaps by the HAR model and comparing it to routines of the resident.
\edit{We use the `Segmentation Accuracy' (Seg. Accuracy) as metric for evaluating our system, where $AC_{identified}$ is the number of action units identified in a given activity; whereas $AC_{activity}$ is the total number of action units that make up a given activity and N is the total number of activity instances:
\begin{equation}
    Seg.\ Accuracy = \dfrac{\sum_{n=1}^{N} (AU_{identified}  \in  AU_{activity})}{\sum_{n=1}^{N} AU_{activity}}
\end{equation}
}
\textit{Evaluating updates:}
The evaluation procedure we use in this work makes use of most recent model obtained so far on all test sets observed after the given update. 
For example, the activity model from train 1 is evaluated on all four test sets observed in the future, whereas the activity model obtained after train 3 update is evaluated on test data observed after said update -- test 3 and test 4 -- both depicted as `M3' and `M4' respectively.
We provide an example of the procedure for the CASAS-Aruba dataset, that spans over 32 weeks of data collection.
During iteration 1, the HAR model trained over Weeks 1-4 (M1) is then evaluated on Weeks 5-6. 
For iteration 2, the HAR model is trained over Weeks 1-6. 
Following evaluation protocol 1, both models (M1 and M2) developed during Weeks 1-4 and Weeks 1-6 are evaluated on Weeks 7-8. 
The procedure is repeated for as long as data is observed in the smart home.

Results for CASAS-Cairo, CASAS-Aruba and CASAS-Milan are reported in Tab. \ref{tab:results-ar-cairo}, Tab. \ref{tab:results-ar-aruba} and Tab. \ref{tab:results-ar-milan} respectively. 
The prominent activities identified for the CASAS-Aruba dataset correspond to Sleep, Work, Meal\_Preparation and Relax. 
For CASAS-Milan, these correspond to Sleep, Kitchen\_Activity (KA), Guest\_Bathroom (GB), Read, Master\_Bedroom\_Activity (MBA), Master\_Bathroom (MB), Watch\_TV, Desk\_Activity (DA), Dining\_Rm\_Activity (DRA).
We observe that with each update the segmentation accuracy for the prominent activities in these three smart homes increases (as depicted in bold text).
Although, the analysis for CASAS-Milan and CASAS-Cairo is over the entire duration of data collected in these homes, we limit the analysis for the CASAS-Aruba dataset to three iterations to showcase the validity of our approach.

CASAS-Aruba provides for a simpler analysis problem, as observed by the relatively high segmentation scores for the prominent activities. 
CASAS-Milan is a complex analysis problem, where the movement of the pet recorded in the home interferes with the movement of the resident, leading to noisy annotations. 

\begin{table}
    \vspace*{0.8em}
    \caption{Experimental Evaluation Cairo:  GT: Ground Truth; M1: Model 1; M2: Model 2. Seg. Accuracy is Segmentation Accuracy for prominent activities.}
    \label{tab:results-ar-cairo}
    \small
    \begin{tabular}{p{2.3cm}|p{2.5cm}|p{2.5cm}} 
        \textbf{Activity} & 
        \text{Test  1} \newline \text{Seg. Accuracy} & 
        \text{Test 2} \newline \text{Seg. Accuracy} 
       
        \\ [10pt]
        \hline
        Eat & GT: 34.96 \textpm 14.63  \newline  M1: 5.86 \textpm 4.72   &  GT: 37.83 \textpm 17.26 \newline M1: 9.72 \textpm 7.67  \newline \textbf{M2: 13.75 \textpm 9.14 }\\ [10pt]
        \hline
        Sleep  & GT: 8.36 \textpm 6.84 \newline M1: 0.0 \textpm 0.0 &  GT: 6.58 \textpm \ 4.44 \newline M1: 0.0 \textpm 0.0  \newline \textbf{M2: 0.291 \textpm 0.99} \\ [10pt]
        \hline
        Work  & GT: 4.94 \textpm 3.50 M1: 0.0 \textpm 0.0 &  GT: 6.75 \textpm 7.39 \newline M1:0.0 \textpm 0.0 \newline \textbf{M2: 0.571 \textpm 1.89} \\ [10pt]
        \hline
        \end{tabular}
\end{table}

\begin{table}
    \vspace*{0.8em}
    \caption{Experimental Evaluation Aruba: GT: Ground Truth; M1: Model 1; M2: Model 2; M3: Model 3. Seg. Accuracy is Segmentation Accuracy for prominent activities.}
    \label{tab:results-ar-aruba}
    \small
    \begin{tabular}{p{1.3cm}|p{2.0cm}|p{2.0cm}|p{2.0cm}} 
        \textbf{Activity} & 
        \text{Test  1} \newline \text{Seg. Accuracy} & 
        \text{Test 2} \newline \text{Seg. Accuracy} &
        \text{Test 3} \newline \text{Seg. Accuracy} 
        \\ [10pt]
        \hline
        Meal\_Prep  & GT: 9.37 \textpm 8.09 \newline M1: 5.11 \textpm 6.65  &  GT: 13.12 \textpm 16.89 \newline M1: 5.61 \textpm 10.01 \newline \textbf{M2: 7.37 \textpm 11.99} & GT: 11.04 
 \textpm 9.68 \newline  M1: 5.18 \textpm 7.10 \newline M2: 4.66 \textpm 7.11 \newline \textbf{M3:6.89 \textpm 6.07} \\ [10pt]
        
        \hline
         Relax  & GT: 3.86 \textpm 3.56 \newline M1: 2.86 \textpm 3.55 &  GT: 4.22 \textpm 3.68 \newline M1: 2.92 \textpm 3.66 \newline \textbf{M2: 3.12  \textpm 3.64}  & GT: 7.12 \textpm 8.07 \newline  M1: 4.24 \textpm 6.60 \newline M2: 3.96 \textpm 6.08 \newline \textbf{M3: 4.45 \textpm 7.03} \\ [10pt]
        
        \hline
        Sleep  & GT: 6.94 \textpm 6.92 \newline M1: 5.10 \textpm 3.75 &  GT: 3.72 \textpm 2.64 \newline M1: 2.59 \textpm 2.87 \newline \textbf{M2: 2.90 \textpm 2.95} & GT: 7.05 \textpm 7.44\newline  M1: 4.11 \textpm 3.35  \newline M2: 5.58 \textpm 5.31 \newline \textbf{M3: 5.58 \textpm 5.31} \\ [10pt]
        
        \hline
        Work  & GT: 3.0 \textpm 3.31 \newline M1: 2.0 \textpm 3.65 &  GT: 3.0 \textpm 1.60 \newline M1: 2.25 \textpm 1.38 \newline \textbf{M2:  2.25 \textpm 1.38} & GT: 6.5 \textpm 7.77 \newline  M1: 6.5\textpm 7.77  \newline M2: 6.5\textpm 7.77 \newline \textbf{M3:  6.5\textpm 7.77}\\ [10pt]
        \hline
        \end{tabular}
\end{table}

\begin{table}
    \vspace*{0.8em}
    \caption{Experimental Evaluation Milan: GT: Ground Truth; M1: Model 1; M2: Model 2 Seg. Accuracy is Segmentation Accuracy for prominent activities.}
    \label{tab:results-ar-milan}
    \small
    \begin{tabular}{p{1.3cm}|p{2.0cm}|p{2.0cm}| p{2.0cm}} 
        \textbf{Activity} & 
        \text{Test  1} \newline \text{Seg. Accuracy} &
        \text{Test 2} \newline \text{Seg. Accuracy} &
         \text{Test 3} \newline \text{Seg. Accuracy}
        \\ [10pt]
        \hline
        Sleep  & GT: 7.05 \textpm 4.50 \newline M1: 0.05 \textpm 0.24  
        & GT: 8.03 \textpm 7.22 \newline M1: 0.0 \textpm 0.0 \newline \textbf{M2: 1.37 \textpm 1.54 }
        & GT: 10.61 \textpm 8.06 \newline M1: 0.0 \textpm 0.0 \newline M2: 0.46 \textpm 1.19 \newline \textbf{M3: 1.84 \textpm 2.64}
        \\[10pt]
        \hline
        KA  & GT: 8.99 \textpm 10.34 \newline M1:7.67 \textpm 9.93  
        & GT: 8.44\textpm 10.05 \newline M1: 0.11 \textpm 0.52  \newline \textbf{M2: 1.14 \textpm 3.71  }
        & GT: 8.90 \textpm 8.93 \newline M1: 0.23 \textpm 0.83  \newline M2:1.36 \textpm 2.90 \newline \textbf{M3: 2.03 \textpm 3.57}\\ [10pt]
        \hline
        GB  & GT: 2.0 \textpm 1.44 \newline M1: 0.0 \textpm 0.0 
        &  GT:1.67 \textpm 1.18 \newline M1: 0.0 \textpm 0.0  \newline \textbf{M2: 0.10 \textpm 0.41 }
        & GT: 2.43 \textpm 1.97 \newline M1:  0.0\textpm 0.0 \newline M2: 0.0 \textpm 0.0 \newline \textbf{M3: 0.16 \textpm 0.55}
        \\ [10pt]
        \hline

        Read  & GT: 7.97 \textpm 6.13 \newline M1: 6.13 \textpm 4.01 

        &  GT: 6.56 \textpm 5.67 \newline M1: 0.07 \textpm 0.50 \newline \textbf{ M2:  3.41 \textpm 3.71}

        & GT: 5.94 \textpm 5.09 \newline M1: 0.08  \textpm 0.36  \newline M2:0.083 \textpm 0.36 \newline \textbf{M3: 2.57 \textpm 3.18}
        
        \\ [10pt]
        \hline
        MBA  & GT: 8.16 \textpm 7.19 \newline M1: 0.32 \textpm 0.80  
        & GT: 6.15 \textpm 5.37 \newline M1: 0.0 \textpm 0.0  \newline \textbf{M2: 1.01 \textpm 1.70}
        & GT: 90 \textpm 9.87 \newline M1: 0.0 \textpm 0.0  \newline M2: 0.0 \textpm 0.0 \newline
        \textbf{M3: 2.36 \textpm 4.36 }
        \\[10pt]
        \hline
        MBath  & GT: 2.78 \textpm 2.31 \newline M1: 0.14 \textpm 0.52  
        & GT: 2.19 \textpm 1.52 \newline M1: 0.0 \textpm 0.0  \newline M2: 0.0 \textpm 0.0 
        & GT: 3.07 \textpm 2.97 \newline M1:  0.0 \textpm 0.0 \newline M2:  0.0 \textpm 0.0 \newline \textbf{M3: 0.20 \textpm 0.59} \\ [10pt]
        \hline
        Watch\_TV  & GT: 9.02 \textpm 6.822 \newline M1: 0.07 \textpm 0.48 
        & GT: 7.51 \textpm 5.67 \newline M1: 0.0 \textpm 0.0  \newline \textbf{M2: 2.81 \textpm 3.54 }
        & GT: 6.38 \textpm 6.37 \newline M1: 0.0 \textpm 0.0 \newline M2: 2.07 \textpm 2.53  \newline \textbf{M3: 2.0 \textpm 2.70} \\ [10pt]
        \hline
        DA & GT: 3.25 \textpm 1.5 \newline M1: 0.0 \textpm 0.0 
        &  GT: 5.9 \textpm 5.52 \newline  M1: 0.0 \textpm 0.0  \newline  M2: 0.0 \textpm 0.0: 
        & GT: 8.82 \textpm 12.64 \newline M1: 0.0 \textpm 0.0  \newline M2:  0.0 \textpm 0.0 \newline \textbf{M3: 3.29 \textpm 3.86}\\ [10pt]
        \hline 
        DRA & GT: 0.0 \textpm 0.0 \textpm 0.0 \newline M1: 0.0 \textpm 0.0  
        & GT: 7.41 \textpm 5.33 \newline M1: 0.0 \textpm 0.0  \newline M2:  0.0 \textpm 0.0 
        & GT: 5.28 \textpm 3.77  \newline M1: 0.0\textpm 0.0  \newline M2: 0.0 \textpm 0.0  \newline \textbf{M3: 0.57 \textpm 0.97}
        \\ [10pt]
        \hline 
        \end{tabular}
\end{table}

\section{Discussion}
\label{sec:Discussion}
\noindent
The premise of this work is that each smart home is different, and--crucially--each resident of such smart homes is different with regard to the activities they engage in. 
Based on these observations (which are backed up, e.g., by the substantial variability in existing smart home datasets), we design the activity recognition system for the home in a fully data-driven approach. 
An initial bootstrapped HAR system is was covered in previous work.
\edit{In this work, we update and extend this recognition procedure, to improve the segmentation accuracy for the identified prominent activities.}

We build on the initial bootstrapped procedure--where the HAR system was initialized \textit{from scratch} and with \textit{minimal supervision}.
Model predictions from this system are used as seed points for developing the update and extend procedure. 
The predictions serve as data points to train the self-supervision module, which is then used to provide predictions at the level of action units.
These predictions in addition to the segmentation provided by the motif models helps improve the segmentation accuracy corresponding to the majority of the activities (that are essential parts of the resident's routines) in the home.
Through an extensive experimental evaluation on three CASAS datasets we demonstrated the effectiveness of the proposed method.
In what follows we discuss additional aspects relevant to our work and outline some next steps for this work.

\subsection{Boosting of Activity Models}
\label{sec:discussion:boosting}
\noindent
Boosting is a method, in the machine-learning sphere, that aims at improving the accuracy of a given learning algorithm  \cite{schapire1999brief, schapire2003boosting, freund1999short, schapire2013boosting}. 
It trains models sequentially by improving learners to provide for a single strong learning model. 
To do so, it assigns higher weights to misclassified data points, such that the subsequent learner labels these data points accurately. 

Our maintenance and update procedure uses similar concepts.
For the updates of the recognition procedure predictions through the bootstrapping procedure are used to train the self-supervision module.
This model then improves segmentation accuracy through providing predictions in the ``non-detection" regions. 
Thus, the overall recognition procedure, at any given time, becomes a ``strong'' learner by using the model predictions from previous iterations.
Contrary to the stopping criteria used in boosting, which stops when the training errors produced by the learner are below a certain given threshold, our update procedure is continuous in identifying activity patterns are observed in the home. 

\subsection{Refinement of Seed Points}
\label{sec:discussion:refinement}
\noindent
Techniques developed for both the bootstrapping and update and extension procedure inform general procedures to develop a functional HAR model in the smart home using minimal supervision from residents.
This recognition procedure gets us closer to a fully functional system capable of identifying the prominent activities as part of the resident's daily living.
Our work thereby serves as proof of concept, which can be further extended and optimized.
We provide predictions corresponding to the ``non-detection" regions, which correspond to portions of data where the initial bootstrapping procedure does not produce any predictions. 
This is achieved through \textit{i)} training a self-supervision based module that uses data and the corresponding predicted label from the initial procedure and, \textit{ii)} using updated motif models. 
Although these serve as good starting points they can be further refined to tune them for specific home settings or activity patterns. 

In both discriminative and generative modeling, posterior probabilities \cite{dietterich2000ensemble,niculescu2005predicting} are used to estimate classification confidence for data points. 
This allows the identification of ``not-so-confident'' data points that can be modeled better. 
Estimating such probabilities is not straightforward when using template-matching-based methods, as in our case of motif models.
Hence the use of additional knowledge may help in identifying confident activity models \cite{smith2012ontology, kim2014ontology, villalonga2017mimu, kang2006wearable} . 
Identification of relevant sensor events, priors on when activities are usually performed in the home, and sequences of activities that make up routines will serve as additional information that can be used to refine the seed points and hence the recognition  that eventually leads to a refined fully functional recognition system. 
It also aids in asking the resident appropriate questions about nuanced activities when they occur in the home. 

\vspace*{-0.5em}
\subsection{Knowledge-based Active Learning}
\label{sec:discussion: knowledge}
\noindent
The methods underlying the HAR system rely on small amounts of activity labels, requested from the resident.
For the initial bootstrapping procedure the focus was on obtaining labels for the most prominent activities (through their motifs), whereas in the update and extension procedure we aim to extend the segmentation boundaries corresponding to these identified activities. 
This developed HAR system proves useful to model the major activities that are part of routines in the home.

Beyond that, for example for very short duration activities, additional methods will be required.
This additional information could be based on heuristics such as using absolute timestamps (or coarse categories like morning or evening sub-routines) and the durations between consecutive analysis segments.
An incremental knowledge-gathering procedure would ensure that we first start from a conservative procedure and move towards a more refined system, in the absence of which we may develop greedy or opportunistic model procedures. 
To incrementally refine and acquire more knowledge about the smart home and its resident, an ontology-based active learning procedure can be employed.
For example, in Milan, two of the activities--`Master\_Bedroom\_Activity' and `Sleep'--are both activities that occur in the bedroom and comprise similar movement patterns. 
Similarly, for activities of `Respirate' / `Meditate' that are predicted as `Work' activities in the bootstrapping procedure and hence not available for analysis in the update and extension procedure. 
A knowledge-based query procedure may help in distinguishing between these closely related yet different activities. 

\subsection{Next Steps for HAR in Smart Home: Routine Assessment}
\label{sec:discussion: anomaly-detection}
\noindent
The activity recognition system developed so far aims at developing a functional HAR system that recognizes regular activities in the home. 
In the next step, we aim to analyze activity routines. 
Routine assessment refers to the problem of identifying regular occurrences of activity sequences in the home \cite{banovic2016modeling}.
Identifying such daily activity routines aids in improving the underlying recognition model through capture of infrequent and short duration activities -- those that are not picked up by current recognition procedures.

The system can aim to provide a trigger to residents or caregivers on identifying anomalies in daily routines thus proving beneficial in assisting living \cite{zhu2015wearable, mandaric2019anomaly, parada2013unsupervised}.
\edit{An example of such an anomalous activity would be fall detection \cite{aran2016anomaly, bakar2016activity, hoque2015holmes, yahaya2019consensus,yoshida2022data}.}

\section{Conclusion}
\label{sec:Conclusion}
\noindent
With reduced sensor costs and advancements in IoT technologies, there is an increased interest in instrumenting ``regular'' homes with sensors that can be used for activity monitoring, turning them into ``smart homes'', which are of benefit, for example, for health care applications. 
However, developing such an activity recognition system is challenging because it needs to be tailored towards the particular smart home and especially towards the resident it is serving.
The overarching goal of our work is to develop methods for automatically deriving tailored HAR systems thereby minimizing user involvement and focusing on the rapid availability of functional HAR systems that are then continuously updated and extended, mainly to capture  activity sequences as shown in this work.

Based on previous work that covers the initial bootstrapping of such a tailored HAR system, in this work, we focused on update and extension of the initial system.
With the proof-of-concept presented, and extensively evaluated, in this paper, the next steps shall focus on capturing less frequently occurring, yet important activities.
Our framework serves as the basis for such an extension and, as such, has the potential to become the overarching, yet extendable, modeling framework for HAR in smart homes.

\section*{Acknowledgment}
This work was partially supported by KDDI Research.
We thank the whole CASAS team and in particular Dr. Diane Cook for aiding us in the process of understanding the CASAS datasets and for providing us with the detailed smart home layouts. We are also grateful for the comments and suggestions provided by the anonymous reviewers who helped us in improving the clarity of the presentation.
\bibliographystyle{IEEEtran}
\bibliography{references.bib}
\end{document}